\documentclass[fleqn,preprint,10pt]{elsarticle}
\usepackage[margin=1.5cm]{geometry}
\usepackage[utf8]{inputenc}
\usepackage[T1]{fontenc}

\usepackage{lineno,hyperref}
\usepackage{amsmath,graphicx,makecell}
\usepackage{changes}
\usepackage{enumitem}
\graphicspath{{./figs/}}
\usepackage{threeparttable}
\usepackage{booktabs}
\usepackage{algorithmic}
\usepackage[linesnumbered, ruled, vlined]{algorithm2e}
\usepackage{footnote}
\usepackage{caption}
\usepackage{subcaption}

\begin{document}
% \flushbottom
% \maketitle
% \thispagestyle{empty}

% \section*{Abstract}
% The operation of the electric grid is undergoing a paradigm shift of transitioning towards carbon neutrality. This perspective article focuses on examining the potential of artificial intelligence (AI) in accelerating the carbon-neutral transition of the electricity sector while, importantly, maintaining reliability of service and cost-affordability. It points out that in order to unlock AI's potential to do so, the AI algorithms originally developed for other applications should be tailored from the perspectives of technology, markets, and policy.

\begin{frontmatter}
\title{Energy System Digitization in the Era of AI: A Three-Layered Approach towards Carbon Neutrality}

\author[1,10]{Le~Xie\corref{cor1}}
\author[2]{Tong~Huang}
\author[1]{Xiangtian~Zheng}
\author[3]{Yan~Liu}
\author[4,5]{Mengdi~Wang}
\author[6]{Vijay~Vittal}
\author[1]{P. R. Kumar}
\author[1]{Srinivas Shakkottai}
\author[7]{Yi~Cui}
\cortext[cor1]{Corresponding author \& lead contact: le.xie@tamu.edu}
\affiliation[1]{organization={Department of Electrical and Computer Engineering},
addressline={Texas A\&M University}, 
city={College Station},
state={TX},
postcode={77843}, 
country={USA}}

\affiliation[2]{organization={Department of Electrical and Computer Engineering},
addressline={San Diego State University}, 
city={San Diego},
state={CA},
postcode={92182}, 
country={USA}}

\affiliation[3]{organization={Computer Science Department},
addressline={University of Southern California}, 
city={Los Angeles},
state={CA},
postcode={90007}, 
country={USA}}

\affiliation[4]{organization={Department of Electrical Engineering and Center for Statistics and Machine Learning},
addressline={Princeton University}, 
city={Princeton},
state={NJ},
postcode={08540}, 
country={USA}}

\affiliation[5]{organization={DeepMind},
city={Mountain View},
state={CA},
postcode={94043}, 
country={USA}}

\affiliation[6]{organization={School of Electrical, Computer and Energy Engineering},
addressline={Arizona State University}, 
city={Tempe},
state={AZ},
postcode={85281}, 
country={USA}}

\affiliation[7]{organization={Department of Materials Science and Engineering},
addressline={Stanford University}, 
city={Stanford},
state={CA},
postcode={94305}, 
country={USA}}

% \affiliation[*]{organization={Corresponding author \& lead contact},
% addressline={le.xie@tamu.edu}}

\begin{abstract} 
The transition towards carbon-neutral electricity is one of the biggest game changers in  addressing climate change since it addresses the dual challenges of removing carbon emissions from the two largest sectors of emitters: electricity and transportation. 
The transition to a carbon-neutral electric grid poses significant challenges to conventional paradigms of modern grid planning and operation. Much of the challenge arises from the scale of the decision making and the uncertainty associated with the energy supply and demand. Artificial Intelligence (AI) could potentially have a transformative impact on accelerating the speed and scale of carbon-neutral transition, as many decision making processes in the power grid can be cast as classic, though challenging, machine learning tasks.
We point out that to amplify AI's impact on carbon-neutral transition of the electric energy systems, the AI algorithms originally developed for other applications should be tailored in three layers of technology, markets, and policy.
\end{abstract}


\end{frontmatter}

\section*{Introduction}
To grapple with climate change, many countries are striving to achieve carbon-neutrality of their electricity sectors. As an example, the U.S. aims for $100\%$ electricity generation from zero-carbon resources by 2035. However, today's decarbonization rate in the U.S. electricity sector may not be able to realize such an aggressive agenda  sector~\cite{EIA2021}.

Speeding up the carbon-neutral transition of the electricity sector requires massive integration of renewable generation at an unprecedented rate. Such large-scale renewable integration poses significant challenges to the operational paradigm that today's grid employs. Reliable operation depends on both offline planning as well as real-time decision making, including monitoring, control, and protection. In the emerging electricity landscape with deep renewables, due to the temporally variable nature of renewable generation, the number of scenarios needing to be considered in the planning studies is extremely large, which creates significant difficulties for grid planners. Moreover, impactful anomalies in a low-carbon grid, e.g., oscillations, and large voltage deviations, may appear more often than in the conventional grid. Consequently, there is a stringent need to monitor and correct these anomalies in a timely manner. Such anomaly monitoring and correction require closed-loop decision-making tools that can convert high-dimensional streaming data into reliable decisions, and apply the decisions to the physical infrastructure in a timely fashion. However, there is generally a lack of such tools for most such anomalous scenarios.

The above operational challenges are likely to become a major bottleneck to accelerating the carbon-neutral transition of the electricity sector. These challenges may benefit from Artificial Intelligence (AI) based solutions. In recent years, AI-based applications are transforming all aspects of human society and endeavor\cite{rolnick2019tackling}. In many successful applications, AI algorithms make decisions based on data without requiring detailed models. This is a highly desirable feature for future power grid operation since accurate physical models of phenomena based on weather or consumer behavior are likely either unknown or complex. On the other hand, as will be seen in what follows, some grid operational needs can be translated into classic AI problems.
However, this will require the translation of the potential of AI into solutions at full-scale that can empower the ambitious carbon-neutral transition in the electricity grid\cite{9091534,chatterjee2022facilitating}, under the additional requirements of safety, time sensitivity, and interpretability.

\begin{figure}
    \centering
    \includegraphics[width =\textwidth]{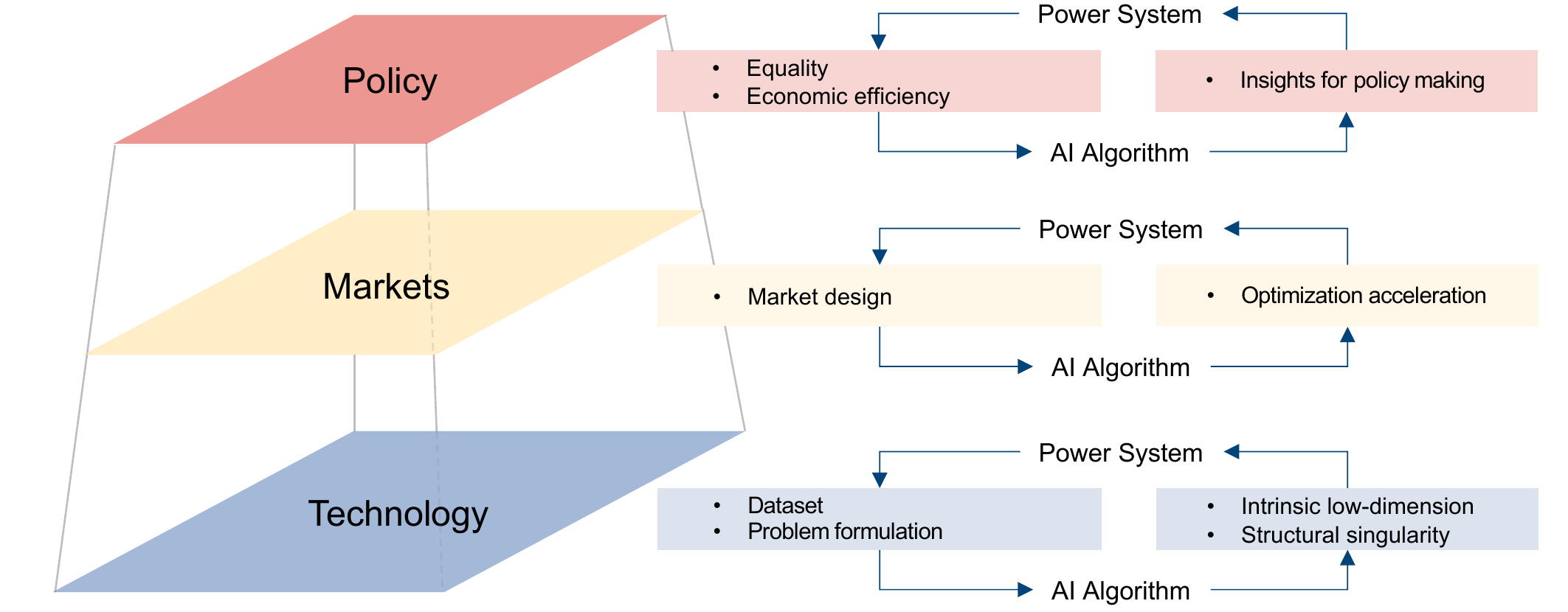}
    \caption{A layered vision of  energy system digitization}
    \label{fig:layered}
\end{figure}

While several survey papers~\cite{rolnick2019tackling, 8649711,xie2022massively} review the applications of AI in electric energy systems and point out the research directions of this field from a technological viewpoint, how to unleash the power of AI in decarbonizing the electricity sector is a complex problem in a broader social-economic-technological space. This Perspective argues that general-purpose AI needs to be carefully tailored in three layers before it can be applied in the safety-critical grid infrastructure applications. These three layers include technology, markets, and policy, as summarized in Figure \ref{fig:layered}.

The rest of this perspective is organized as follows. First, we briefly introduce key decision-making mechanisms associated with the operation of modern power grids. Then we highlight the connection between grid operation and AI. Finally, we elaborate on the tri-layer tailoring of AI for the carbon-neutral transition of the electric grids.

\section*{The Role of AI in Carbon-neutral Transition of Power Grid} 
\begin{figure}
    \centering
    \includegraphics[width =7in]{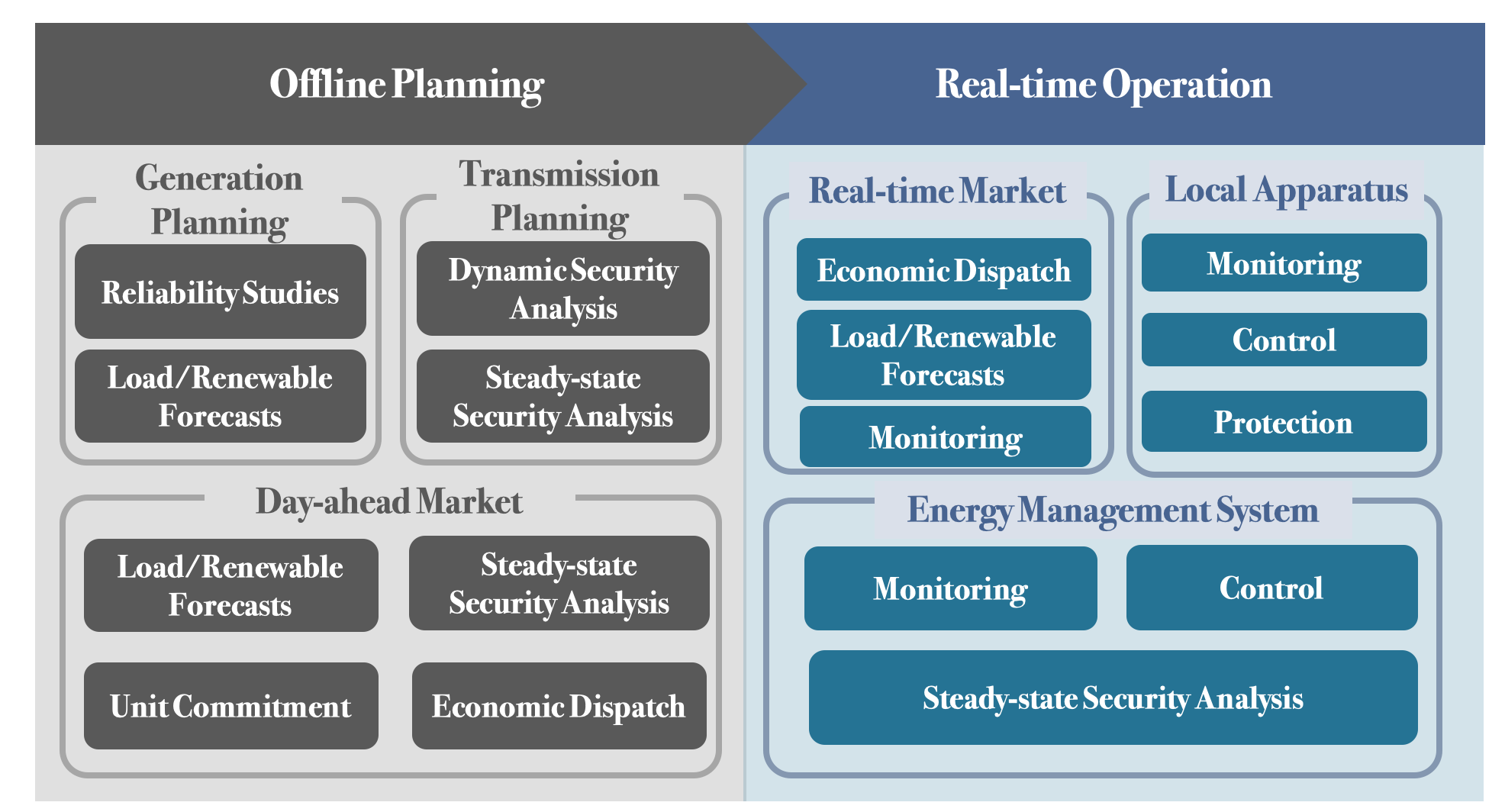}
    \caption{Decision making modules of power (transmission) system operation.}
    \label{fig:power_grid_operations}
\end{figure}

This section elaborates on AI's potential for addressing challenges in the carbon-neutral transition of the electricity energy systems. We provide a brief overview of power grid operation (see a detailed overview of power system basic functionalities in the review paper~\cite{xie2022massively}) and highlight the decision making processes (Figure~\ref{fig:power_grid_operations}). Then we present the connection between these decision-making processes and representative problems in the AI field.

\subsection*{Overview of Modern Power Grid Operation}

The stringent balance requirement between electricity supply and demand in the modern grid is accomplished by a two-stage strategy which consists of offline planning and real-time operation\footnote{The grid operation described here is for transmission systems. The decision-making processes in current distribution system operation generally are a subset of those in transmission systems.}. \underline{Offline planning} includes generation and transmission planning and several forward financial markets. \underline{Generation planning} leverages \underline{reliability studies} to decide when and how much additional generation capacity should be added over a planning horizon of several years \cite{bebic2008power}. The objective of \underline{transmission planning} is to ensure that the transmission lines deliver electricity from the generators to the loads. In the transmission planning stage, system planners scrutinize the grid behavior at multiple time scales by comprehensively conducting physical model-based simulations for future possible scenarios based on past operational experience. Such a procedure is called \underline{system security analysis}. Once the power grid is physically established, system operators can leverage financial markets, such as day-ahead, reserve, and capacity markets, for enhancing the grid's economical efficiency. \underline{Unit commitment} (UC) and \underline{economic dispatch} (ED) are key modules that support energy trading in a \underline{day-ahead market}.

The decision making in \underline{real-time operation} leverages real-time measurements \cite{sztipanovits2012model}, e.g., those obtained over SCADA (supervisory control and data acquisition), and PMUs (phasor measurement units), in a hierarchical manner. At the global level, the energy management system (EMS) at a control center pre-processes raw measurements from the grid and provides system operators with \underline{monitoring} and \underline{control} functionalities. System security analysis is also conducted based on the most recent measurements. At the local level, grid components are regulated and protected by local control and \underline{protection} apparatuses. The economic efficiency of the grid is ensured by a real-time (spot) market. The spot electricity prices are determined by solving ED problems based on the most recent measurements and short-term load forecasts.

\subsection*{Background of Grid Operation with AI Support}
This subsection establishes connections between the decision making
modules of grid operation and the three representative formulations of AI problems, i.e., supervised, unsupervised, and
reinforcement learning.

\subsubsection*{Supervised Learning for Prediction and Detection}
Given a set of vectors with labels, a supervised learning method is used for constructing the relationship between the vectors and their labels. Formally, suppose that a procedure takes vector $\mathbf{X}$ as input and generates a label vector $\mathbf{Y}$, where the label can be either numerical or categorical. The elements of $\mathbf{X}$ are called ``features." The supervised learning method aims to construct a function $f$ that maps $\mathbf{X}$ to $\mathbf{Y}$, by learning from $N$ historical records of input-output pairs generated from the procedure, i.e., $\{(\mathbf{x}_1, \mathbf{y}_1),(\mathbf{x}_2, \mathbf{y}_2),\ldots, (\mathbf{x}_N, \mathbf{y}_N)\}$. When a new input, say, $\mathbf{x}_{N+1}$, appears,  the resulting $\mathbf{y}_{N+1}$ is expected to be approximated by $f(\mathbf{x}_{N+1})$ with reasonable accuracy. If the number of options for $\mathbf{Y}$ is finite, the problem is called one of \emph{classification}; if continuous variables constitute the output vector $\mathbf{Y}$, the problem is called \emph{regression}\cite{bishop2006pattern}.

Many decision making modules for grid operation involve classification and regression with clear definitions of the input vector $\mathbf{X}$ and the response vector $\mathbf{Y}$. For example, it can be shown that battery cycle life can be predicted by linear regression with properly selected input features\cite{severson2019data}. Another example concerns grid online stability assessment that aims to determine if the system is stable given real-time measurements. This is a standard classification problem, with the measurements as inputs and a binary variable (1: stable or 0: unstable) as output. Conventional stability assessment relies heavily on time-consuming simulations that cannot be carried out in real time. It has been reported that a Convolution Neural Network (CNN) can quickly achieve high accuracy\cite{8920121}. Table \ref{Tab:Supervised_learning} shows other AI adoption examples with inputs and outputs clearly defined in the context of power grids.
Researchers are striving to build state-of-the-art machine learning benchmarks for several challenging use cases with open datasets~\cite{zheng2021psml}.
\begin{table}[h]
\caption{Power Grid Applications with Supervised Learning Formulation}
\label{Tab:Supervised_learning}
\begin{tabular}{l|l|l|l}
\hline
\hline
\textbf{Challenge}           & \textbf{Application} & \textbf{Input}: $X$ & \textbf{Output}: $Y$ \\ \hline
\begin{tabular}[c]{@{}l@{}}Load/Renewable \\Forecasts \end{tabular}      & \begin{tabular}[c]{@{}l@{}}Short-term forecast\cite{6945846}\end{tabular}&\begin{tabular}[c]{@{}l@{}}Data on energy\\ and/or weather\cite{6945846} \end{tabular}&   \begin{tabular}[c]{@{}l@{}} PV forecasts\cite{6945846}\end{tabular}\\ \hline

Reliability Studies & \begin{tabular}[c]{@{}l@{}}
Fast state evaluation\cite{8514049}\\state enumeration \cite{9136915}\end{tabular}& \begin{tabular}[c]{@{}l@{}}Bus generation\\info. \cite{8514049}, weather\cite{9136915}\end{tabular}&\begin{tabular}[c]{@{}l@{}}Success/failure\\status of buses\cite{8514049},\\renewable generation \cite{9136915}\end{tabular}\\ \hline

Security Analysis   &  \begin{tabular}[c]{@{}l@{}}External grid modeling\cite{6152193}, \\ contingency screening \cite{8705389} \end{tabular}&\begin{tabular}[c]{@{}l@{}}Interface voltage \cite{6152193},\\bus flow injection \cite{8705389}\end{tabular}                &\begin{tabular}[c]{@{}l@{}}Interface power\\ flow \cite{6152193},\\ bus voltage and \\security index\cite{8705389}\\\end{tabular} \\ \hline
\begin{tabular}[c]{@{}l@{}} Unit Commitment \& \\ Economic Dispatch \end{tabular}            & \begin{tabular}[c]{@{}l@{}}%SCED\cite{9177087,9426454,9479716,9392288,9069289,9115822,8447254,8802273,9244070,8693983}\\
Binding constraint\\prediction \cite{9388933},
constrain \\convex relaxation\cite{9340012}\end{tabular}& \begin{tabular}[c]{@{}l@{}} Scenarios \cite{9388933}\\3-phase voltage\cite{9340012}\\\end{tabular}& \begin{tabular}[c]{@{}l@{}} Binary variable\\for each constrain \cite{9388933},\\power flow \cite{9340012}\\\end{tabular}\\\hline

Monitoring          &\begin{tabular}[c]{@{}l@{}}Anomaly classification\cite{8265216},\\stability assessment\cite{8920121},\\ battery cycle \\life prediction\cite{severson2019data}\end{tabular}&\begin{tabular}[c]{@{}l@{}}Streaming data, \cite{8265216,8920121}\\info. from  discharge \\voltage curves and \\ capacityfade curves\cite{severson2019data}\end{tabular}   &\begin{tabular}[c]{@{}l@{}}Binary indicator \cite{8920121},\\ event type\cite{8265216},\\battary cycle life\cite{severson2019data}\end{tabular} 
\\ \hline

\hline
\hline
\end{tabular}
\end{table}

\begin{table}[tbh]
\caption{Power Grid Applications with Unsupervised Learning Formulation}
\label{Tab:Unsupervised_learning}
\begin{tabular}{l|l|l|l}
\hline
\hline
\textbf{Challenge}           & \textbf{Application} & \textbf{Samples} & \textbf{Goal}\\ \hline
\begin{tabular}[c]{@{}l@{}}Load/Renewable \\Forecasts \end{tabular}      & \begin{tabular}[c]{@{}l@{}}Scenario generation \cite{8260947}\end{tabular}&\begin{tabular}[c]{@{}l@{}}Data of energy \\and weather\cite{8260947} \end{tabular}&   \begin{tabular}[c]{@{}l@{}}Density estimation\cite{8260947}\end{tabular}\\ \hline

Security Analysis   &  \begin{tabular}[c]{@{}l@{}}Load modeling\cite{4282059}\\Generator coherency\\identification\cite{7243366} \end{tabular}&\begin{tabular}[c]{@{}l@{}}Customer energy data \cite{4282059},\\synchrophasor data \cite{7243366}\end{tabular}                &\begin{tabular}[c]{@{}l@{}}Clustering\cite{4282059,7243366}\end{tabular}           \\ \hline

Monitoring          &\begin{tabular}[c]{@{}l@{}}Anomaly detection\cite{6808416}\\localization\cite{9043670}\end{tabular}&\begin{tabular}[c]{@{}l@{}}Streaming data \cite{6808416,9043670}\end{tabular}   &\begin{tabular}[c]{@{}l@{}}Dimensionality\\ reduction\cite{6808416,9043670} \end{tabular}           \\ \hline
\hline
\hline
\end{tabular}
\end{table}

\subsubsection*{Unsupervised Learning for Modeling High-dimensional Data}
Unsupervised learning addresses samples without labels. Suppose that only $N$ samples $\{\mathbf{x}_1, \mathbf{x}_2,\ldots, \mathbf{x}_N\}$ are available, without any labels. Unsupervised learning aims to achieve one of the following objectives: 1) \underline{Clustering} aims to find samples that share similarities; 2) \underline{Density estimation} aims to determine the probability distribution governing the given samples (explicitly or implicitly), and to generate new samples with the same probability distribution as the given samples; and 3) \underline{Dimensionality reduction} attempts to project the high-dimensional samples into a low-dimensional space that allows for visualization or easily discovering irregularities in the samples.

The three objectives above are relevant in addressing some challenges associated with the grid. For example, the offline planning problem requires realistic scenarios, including load data conditioned on weather and the season. Generating the scenarios can be formulated as a density estimation problem which can then generate new scenarios with the same probability density as historic scenarios. The density estimation problem can be solved by a Generative Adversarial Network (GAN)\cite{8260947}. In a distribution system, a utility company divides its customers into several representative clusters in order to understand customers' behaviors \cite{4282059}. This problem can be solved by various clustering algorithms, such as k-means\cite{4282059}. In addition, decision making in the control center often involves high-dimensional data collected from a wide area of the grid. It can be shown that the high-dimensional grid data can be projected into a low-dimensional space by principal component analysis (PCA), allowing for efficient data storage and early event detection\cite{6808416}. Table \ref{Tab:Unsupervised_learning} maps some power system challenges to the three goals of the unsupervised learning.

\subsubsection*{Reinforcement Learning}
The goal of reinforcement learning (RL) is to make decisions by interacting with an environment under study\cite{bishop2006pattern}. Generally, a RL algorithm includes the following elements: states, actions, a policy, and a reward function. A state $\mathbf{s}_k$ at time $k$ can be measured from the environment, and an action $\mathbf{a}_k$ can be applied to the environment at time $k$, resulting in a new state $\mathbf{s}_{k+1}$. The state transition from $\mathbf{s}_k$ to $\mathbf{s}_{k+1}$ is determined by the environment whose governing laws are either unknown or complex. The action and the resulting state transition lead to a real-value reward assigned by a reward function $R(\mathbf{s}_k,\mathbf{a}_k, \mathbf{s}_{k+1})$. A policy $\pi$ maps a state to an action. RL searches for an optimal policy that maximizes the total reward over a planning horizon. Possible applications of RL in power systems include model calibration\cite{9220820}, demand modeling\cite{9068279}, energy trading\cite{8662721}, reactive power control\cite{1318666}, tap changer control\cite{8873679}, and relay timer setting in distribution systems\cite{9029268}. Table \ref{tab:RL} identifies the standard RL ingredients for each of the algorithms.

\begin{table}[h]
\caption{{Power Grid Applications with Reinforcement Learning Formulation}}
\label{tab:RL}
\begin{tabular}{l|l|l|l|l}
\hline
\hline
\textbf{Challenge} & \textbf{Application}                                                             & \textbf{State $\mathbf{s}_k$}& \textbf{Action $\mathbf{a}_k$}& \textbf{Policy Goal $R$} \\ \hline
\begin{tabular}[c]{@{}l@{}}Security\\Analysis\end{tabular}  & \begin{tabular}[c]{@{}l@{}}Model calibration\cite{9220820},\\ Demand modeling\cite{9068279}\end{tabular}     & \begin{tabular}[c]{@{}l@{}}Model parameters\cite{9220820},\\load fraction\cite{9068279}\end{tabular}                                &\begin{tabular}[c]{@{}l@{}}model parameter\\ modification\cite{9220820}\\load fraction\\ modification \cite{9068279}\end{tabular}                 & \begin{tabular}[c]{@{}l@{}}Response \\mismatch\\ minimi\\-zation\cite{9220820,9068279}\end{tabular}\\ \hline
\begin{tabular}[c]{@{}l@{}} Unit Commitment \& \\ Economic Dispatch \end{tabular}           & Energy trading\cite{8662721}&\begin{tabular}[c]{@{}l@{}}Past wholesale\\price-quantity pairs,\\ \& retail demand-\\price pairs\cite{8662721}\end{tabular}                                &Retail price\cite{8662721} &\begin{tabular}[c]{@{}l@{}}Profit maximi\\-zation of a load  \\serving entity\cite{8662721}\end{tabular}\\ \hline
Control            & \begin{tabular}[c]{@{}l@{}}Reactive power control\cite{1318666},\\ tap changer control\cite{8873679}\end{tabular}   & \begin{tabular}[c]{@{}l@{}}Binary security status\cite{1318666},\\voltage \&tap ratio\cite{8873679}\end{tabular}&\begin{tabular}[c]{@{}l@{}}Tap ratio \cite{1318666,8873679}\\ \&reactive power\\compensation\cite{1318666}\end{tabular}          &\begin{tabular}[c]{@{}l@{}}Flow/voltage\\tracking\cite{1318666,8873679}\end{tabular}\\ \hline

Protection         & \begin{tabular}[c]{@{}l@{}}Relay timer setting in \\ distribution systems\cite{9029268}\end{tabular} & \begin{tabular}[c]{@{}l@{}}Line current,\\ breaker status, \\ \& Timer value\cite{9029268}\end{tabular}&timer setting\cite{9029268}& \begin{tabular}[c]{@{}l@{}}Minimizing \\mis-operation\\rate\cite{9029268}\end{tabular}\\ \hline\hline
\end{tabular}
\end{table}

\section*{A Three-Layered Approach to Tailor Design AI for Carbon-neutral Electric Grids}
\label{sec:AI_advances}

Given the safety criticality, time-sensitivity, and desire for interpretability in the electric power system operation, we postulate a three-layered approach to tailor AI for power system applications, namely Technology, Markets, and Policy. At each layer, the domain-specific constraints require AI development to be suitably designed; innovations in AI also shed light on how each layer can be further evolved by taking full advantage of future developments. 
\subsection*{At the Technology Layer}

Power grid operation entails complicated interactions of millions of physical components.
The decision making processes over the electricity infrastructure typically employ the mathematical descriptions of these interactions, which in turn are derived from first principles, such as Newtonian mechanics and electrodynamics. The mathematical descriptions of these processes may possess some properties that may not be straightforward to discern,  even for power engineers. For example, high-dimensional data concerning electrical variables, such as measurements of voltage and current over the power grid, may possess a low-rank structure. This is because electrical variables at different locations are correlated by transmission/distribution lines. The hidden properties of the nature of the power grid phenomena can guide the selection / design of AI algorithms to address the operational challenges facing power grids. One can consider the localization of the source of the forced oscillation as an example, which consists of the determination the location of anomalous sources given the measurements over the grid. At first glance, the source localization problem seems to be closely related to a typical supervised classification problem, in that we aim to classify the locations into two categories, i.e., locations close to the anomaly source versus locations far away from the source. The performance of the supervised learning algorithm generally depends heavily on the size and data quality of the training sets. However, it is generally challenging to generate a large, high-quality training set for large-scale power systems. By recognizing the low-rank property of the sensor measurements over the grid, and the sparsity property of anomalous sources, one can choose an unsupervised learning algorithm called Robust Principal Component Analysis (RPCA) to pinpoint the source. Such an algorithm does not require a large amount of training data and exhibits promising performance~\cite{9043670}. While it is well accepted that domain knowledge can help identify application scenarios of general-purpose AI algorithms, as well as to select useful features feeding the AI algorithms, the structural properties that are hidden under the complicated mathematical descriptions of the electricity infrastructure should be exploited at all stages of development of the AI algorithms for grid applications, in order to obtain robust, interpretable AI-powered tools for the grid applications.

\subsection*{At the Markets Layer}
One of major AI applications in electricity markets is to accelerate large-scale optimization.
For instance, unit commitment and economic dispatch, which are fundamental problems in electricity markets, face critical challenges due to increasing uncertainty caused by deepening penetration of renewables.
Several data-driven scenario-based optimization approaches have been proposed to efficiently obtain optimal solutions that explicitly provide the probability that the solution is feasible~\cite{geng2019data,modarresi2018scenario,ming2017scenario,xavier2021learning}.
Another machine learning-based approach has been proposed to identify active sets of safety constraints to obtain optimality more efficiently~\cite{misra2022learning}.
Additionally, the advances in AI can potentially help to optimize the allocation of market investment resources. As an example, machine learning-based demand response from just a few targeted locations has been found to be effective in mitigating price volatility. Therefore, resources for demand response programs can be strategically concentrated in targeted areas, rather than one size fits all, to achieve the most effective social welfare improvements~\cite{lee2022targeted}.

Although there have been many studies showing the potential of AI technology in improving the reliability and economy of electric grids, the lack of appropriate market design is holding it back.
For example, a reinforcement learning-based framework has been proposed to provide voltage regulation via reactive power support in distribution grids with deep solar photovoltaic penetration~\cite{el2021fully}. However, the lack of auxiliary markets in distribution grids hinders the realization of economic revenue from providing reactive power support, thus weakening the economic impacts of AI techniques.
Therefore, changes to the design of the electricity market are imperative to accommodate technological innovation and translate advances in AI into reality.

\subsection*{At the Policy Layer}

Electric grids are heavily regulation and policy driven.  With increasing complexity of electric grids, AI ushers in unique opportunities to integrate interdisciplinary knowledge and leverage heterogeneous data to provide effective insights for policy making.
For example, a machine learning-based approach has been proposed to comprehend the correlation between electricity consumption, number of COVID-19 cases, level of social distancing, and degree of commercial activity during the COVID-19 pandemic, using cross-domain datasets. It can be used as an indicator for predicting changes caused by such an unprecedented event~\cite{ruan2020cross}. 
As another example, the design of green energy policies, such as encouraging the adoption of household rooftop solar panels, will require insights from AI techniques. An example is a machine learning-based solar deployment database in the U.S.~\cite{yu2018deepsolar} to improve energy justice and equity for diverse populations, and analyze social impacts on job creation. An AI-informed policy design would be a prudent approach to driving the carbon neutral transition in the electricity sector. On the other hand, new regulations must be developed to regulate AI applications in such a critical infrastructure system to ensure reliability and privacy. For example, an AI application must possess some key properties, such as interpretability, to facilitate further inspection and investigation by human operators. The increasing demand for data acquisition in power grids also requires new regulations on data privacy and data availability and regulate data acquisition processes to protect the privacy of data owners while maximizing the utility of data.

\section*{Concluding Remarks}
\label{sec:concl}
Energy system decarbonization is one of the most challenging and exciting areas of research and innovation for the 21st century. This article presents a perspective on how digitization and AI could play a crucial role in the carbon neutral transition of the energy sector. We argue that higher impact of digitization and AI in the electric energy industry could be achieved through a ``three-layered'' integrated approach that encompasses technology, markets, and policy layers. Domain-tailored digitization and AI will draw upon unique specifications in all three layers in the energy sector, while providing fertile ground for use-inspired innovations in methodology and algorithms. There are several actionable recommendations that would potentially bring the energy and AI communities closer together:

\begin{itemize}
    \item To involve more power domain-agnostic researchers from the 
    broader AI community, the energy and power community should develop a suite of problem formulations that are accessible to general AI researchers. These problems should be motivated by real-world needs and have the potential to engage the AI community. As an example, the workshop on ``Learning to run a power network (L2RPN)'' provides a platform for such problem definition and solution~\cite{L2RPN}. 
    \item A suite of open, cross-domain data sets that are well benchmarked and labeled for representative power system operating conditions should be developed and shared with the broader AI community. A possible example is an open-source cross-domain dataset~\cite{ruan2020cross} that includes electricity consumption, public health, and mobility data, which was released for broader communities to understand the short-run impact of COVID on the U.S. electricity sector in a data-driven manner.
    \item On the educational front, both the power/energy community and the AI/digitization community should provide use-inspired cases and tools that will be accessible to undergraduates in both areas. Helping students become ``bi-lingual'' in both energy and AI terms would be important. Examples include data science and machine learning courses for power systems in universities, such as Texas A\&M University~\cite{course1}, University of Texas, Austin~\cite{course2}, and University of Washington, Seattle~\cite{course3}.
\end{itemize}

\section*{Materials Availability}
This study did not generate new unique reagents.

\section*{Data and Code Availability}
This study does not involve data or code.

\section*{Acknowledgement}
This material is based upon work partially supported by the U.S. Department of Energy's Office of Energy Efficiency and Renewable Energy (EERE) under the Solar Energy Technologies Office Award Number DE-EE0009031, and by the U.S. National Science Foundation under the Grants CMMI-2130945 and ECCS-2038963. The views expressed herein and conclusions contained in this document are those of the authors and should not be interpreted as representing the views or official policies, either expressed or implied, of the Department of Energy,  or the United States Government.

\section*{Author Contributions}
Conceptualization, L.X.;
Methodology, L.X. and T.H.;
Investigation, L.X. and T.H.;
Writing – Original Draft, L.X., T.H., and X.Z.;
Writing – Review \& Editing, L.X., T.H., X.Z., Y.L., M.W., V.V., P.R.K., S.S., and Y.C.;
Funding Acquisition, L.X.;
Supervision, L.X.

\section*{Declaration of Interests}
The authors declare no competing interests.

\Urlmuskip=0mu plus 1mu\relax  % split url
\bibliographystyle{elsarticle-num}
% \bibliography{ref2}

\end{document}